\documentclass[3p,authoryear]{elsarticle}
\makeatletter
\def\ps@pprintTitle{%
  \let\@oddhead\@empty
  \let\@evenhead\@empty
  \def\@oddfoot{\reset@font\hfil\thepage\hfil}
  \let\@evenfoot\@oddfoot
}
\makeatother

\usepackage{graphicx}
\usepackage{subfig}
\usepackage{amsmath, amsfonts, amssymb, bm}
\usepackage{geometry}%
\usepackage{booktabs}

\usepackage{color, xcolor, colortbl}
\usepackage{hyperref}
\hypersetup{
  colorlinks,
  citecolor=blue,
  filecolor=blue,
  linkcolor=blue,
  urlcolor=blue
}

\usepackage{fourier}
\DeclareMathAlphabet{\mathcal}{OT1}{pzc}{m}{it}
\DeclareSymbolFont{letters}{OML}{cmm}{m}{it}

\usepackage{color, xcolor, colortbl}
\usepackage{hyperref}
\hypersetup{
  colorlinks,
  citecolor=blue,
  filecolor=blue,
  linkcolor=blue,
  urlcolor=blue
}

\usepackage{listings}
\definecolor{mygreen}{RGB}{28,172,0} 
\definecolor{mylilas}{RGB}{170,55,241}
\lstdefinestyle{custommatlab}{
  language=Matlab,%
  breaklines=true,%
  basicstyle=\footnotesize,
  morekeywords={matlab2tikz},
  keywordstyle=\color{blue},%
  morekeywords=[2]{1}, keywordstyle=[2]{\color{black}},
  identifierstyle=\color{black},%
  stringstyle=\color{mylilas},
  commentstyle=\color{mygreen},%
  showstringspaces=false,
  emph=[1]{for,end,break},emphstyle=[1]\color{red}, 
}
\lstdefinestyle{customc}{
  belowcaptionskip=1\baselineskip,
  breaklines=true,
  language=C++,
  showstringspaces=false,
  basicstyle=\footnotesize\ttfamily,
  keywordstyle=\bfseries\color{green!40!black},
  commentstyle=\itshape\color{purple!40!black},
  identifierstyle=\color{blue},
  stringstyle=\color{orange},
}
\begin{document}
\begin{frontmatter}
  \title{Matlab vs. OpenCV:\\A Comparative Study of Different Machine Learning
    Algorithms\tnoteref{t1}}\tnotetext[t1]{This manuscript was initially composed in 2011 as part of
    a research pursued that time. This paper is currently under consideration in Pattern Recognition
    Letters.}  \author[AAE]{Ahmed A. Elsayed} \ead{AhmedAnis03@gmail.com}

  \address[AAE]{B.Sc., Senior Data Scientist, TeraData, Egypt.}

  \author[WAY]{Waleed~A.~Yousef\corref{cor1}}
  \ead{wyousef@GWU.edu, wyousef@fci.helwan.edu.eg}
  \cortext[cor1]{Corresponding Author}

  \address[WAY]{Ph.D., Computer Science Department, Faculty of Computers and Information, Helwan
    University, Egypt.\\ Human Computer Interaction Laboratory (HCI Lab.), Egypt.}

  \begin{abstract} Scientific Computing relies on executing computer algorithms coded in some
    programming languages. Given a particular available hardware, algorithms speed is a crucial
    factor. There are many scientific computing environments used to code such algorithms. Matlab is
    one of the most tremendously successful and widespread scientific computing environments that is
    rich of toolboxes, libraries, and data visualization tools. OpenCV is a (C++)-based library
    written primarily for Computer Vision and its related areas. This paper presents a comparative
    study using 20 different real datasets to compare the speed of Matlab and OpenCV for some
    Machine Learning algorithms. Although Matlab is more convenient in developing and data
    presentation, OpenCV is much faster in execution, where the speed ratio reaches more than 80 in
    some cases. The best of two worlds can be achieved by exploring using Matlab or similar
    environments to select the most successful algorithm; then, implementing the selected algorithm
    using OpenCV or similar environments to gain a speed factor.
  \end{abstract}

  \begin{keyword}
    Matlab \sep OpenCV \sep Machine Learning.
  \end{keyword}

\end{frontmatter}


\section{Introduction}
\subsection{Background}\label{sec:background}
Scientific Computing relies on executing computer algorithms written in some programming
languages. Given a particular available hardware, algorithms speed is a crucial factor. There are
many scientific computing environments used to code such algorithms. Matlab is a high-level
programming language for scientific computing \citep{MathWorks2011MATLAB}. It is an array-based
programming language, where an array is the basic data element. Matlab has an extensive scientific
library and toolboxes across different areas of science, in addition to its data visualization
capabilities and functionalities. OpenCV \citep{Bradski2000OpenCV} on the other hand, developed by
Intel and now supported by Willow Garage \citep{WillowGarage}, is a free library release under BSD
license. It aims at providing a well optimized, well tested, and open-source (C++)-based
implementation for computer vision algorithms.

\bigskip

Code developing, algorithm implementation, data presentation, and other scientific computing
activities are much easier and more convenient in Matlab or similar environments. However,
algorithms developed in native languages, e.g., C++, execute much faster. In research projects that
deliver final products, in particular machine learning and model selection based projects, one
usually tries several models and run hundreds, or even thousands, of experiments before settling on
the final model. It is much more convenient and time saving then to leverage an environment with a
rich toolboxes and libraries as Matlab. However, the final selected model has to execute fast for
satisfactory performance. This may require recoding the selected model anew using a native language,
e.g., C++.

\subsection{Motivation}\label{sec:motivation}
The motivation behind the present article was a real life project for developing a Computer Aided
Detection (CAD) to detect breast cancer in digital mammograms \citep{Yousef2010OnDetecting,
  AbdelRazek2013MicroclacificationLIBCAD, AbdelRazek2012MicroclacificationLIBCAD,
  Yousef2017MethodSystemForComputer}. Using Matlab was extremely efficient in trying dozens of
machine learning and image processing algorithms. When the project reached the point of deploying
the selected models to a final product it was necessary to write some of those selected algorithms
in plain C to gain a speed factor. Moreover, sometimes it was necessary in the developing phase,
before deployment, to code some of the machine learning algorithms in plain C, to be able to run
thousands of experiments in reasonable time.

We spent significant amount of time during the past years developing rigorous methods for classifier
assessment including nonparametric estimation from one dataset \citep{Yousef2004ComparisonOf,
  Yousef2019AUCSmoothness-arxiv}, assessment in terms of Partial Area Under the ROC Curve (PAUC)
\citep{Yousef2013PAUC}, uncertainty estimation of the performance \citep{Yousef2005EstimatingThe},
uncertainty estimation from two independent sets \citep{Yousef2006AssessClass, Chen2012ClassVar},
uncertainty estimation using cross validation estimator
\citep{Yousef2019EstimatingStandardErrorCross-arxiv, Yousef2019LeisurelyLookVersionsVariants-arxiv,
  Yousef2009EstCVvariability}, and the best practices for classifier design and assessment
\cite{Chen2012UncertEst, Shi2010MAQCII, Yousef2014LearningAlgo,
  Yousef2019PrudenceWhenAssumingNormality-arxiv}, among many others.

However, we learned that all of these successful methodologies may be very difficult to apply if the
learning algorithm is computationally very slow, the matter that compromises their
utility. Therefore, it is not only important to have a fast algorithm for its own sake but also for
the sake of accuracy estimation and assessment, which is mandatory for sound conclusion and
results. Yet, the speed of these learning algorithms is not determined only by their design but also
by their computing environment.

\bigskip

In retrospect, from the introduction above, the present article complements our theoretical work on
assessment and provides a very practical comparative study between Matlab and OpenCV using 20 real
datasets, to compare their execution times for different machine learning algorithms. It is
impressive that in some cases OpenCV is 80 times faster than Matlab.

\subsection{Manuscript Roadmap}\label{sec:manuscript-roadmap}
In Section \ref{SecMethods} we describe the settings of this study including
hardware, software, datasets, and the machine learning algorithms used for comparison. Section
\ref{SecResults} presents the results and discussion.

\section{Methods}\label{SecMethods}
\begin{table}\centering
    \begin{tabular}{ccccc}
      \toprule
      &Data sets &$n$& $p$  & $C$ \\
      \midrule
      \smash{$D_{1}$} & Connectionist\  Bench & 208 & 60 & 2\\

      \smash{$D_{2}$}&Breast\   Tissue & 106 & 9 &4 \\

      \smash{$D_{3}$}&Blood\  Transfusion  & 748 & 4 & 2\\

      \smash{$D_{4}$}& Glass\  Identification  & 214 & 9 & 6 \\

      \smash{$D_{5}$}&Haberman's\  Survival& 306 & 3 & 2\\

      \smash{$D_{6}$}&Image\  segmentation & 210 & 19 & 7\\

      \smash{$D_{7}$}& Iris  &150 & 4 & 3\\

      \smash{$D_{8}$}& MAGIC\  Gamma  & 19020 & 10 & 2 \\

      \smash{$D_{9}$}&shuttle  & 43500 & 9 & 7\\

      \smash{$D_{10}$}& Letter\ Recognition & 20000 & 17 & 26 \\

      \smash{$D_{11}$}& Arcene & 100 & 10000 & 2 \\

      \smash{$D_{12}$}& Madelon  & 2000 & 500 & 2 \\

      \smash{$D_{13}$}& Pen\  Based\ Recog.  & 7494 & 16 & 9 \\

      \smash{$D_{14}$}& SPECT Heart & 80 & 22 & 2 \\

      \smash{$D_{15}$}& SPECTF Heart  & 80 & 44 & 2 \\

      \smash{$D_{16}$}& Arcene Test  &167 & 10000 & 2 \\

      \smash{$D_{17}$}& Madelon Test & 1800 & 500 & 2  \\

      \smash{$D_{18}$}& Pen Based Recog. Test & 3489&16 & 9  \\

      \smash{$D_{19}$}& SPECT Heart Test &187 &22 & 2  \\

      \smash{$D_{20}$}& SPECTTF Heart Test &187 & 44 & 2 \\
      \bottomrule
    \end{tabular}
    \caption{20 UCI real datasets used to compare performance; $n$, $p$, $C$ are \#observations,
      \#features, and \#Classes respectively.}\label{table:Datasets}
  \end{table}
In this section we introduce the environment used in conducting our comparative study. We describe the hardware, software, datasets, and the machine learning algorithms used as a basis for comparison.

\bigskip

We ran all experiments on an Intel core 2 duo P7450 machine, with 3GB RAM, and Ubuntu 11.04 32-bit Operating System. We used Matlab version 7.12.0.635 (R2011a), and OpenCV C++ version 2.1 without \emph{TBB library} \citep{Intel2011TBB}. This is to ensure that we use sequential processing as Matlab, to establish similar environment conditions. Code was compiled using \texttt{gcc} compiler version 4.5.2.

We used 20 different real datasets, obtained from UCI Machine Learning Repository \citep{Blake1998UCIRepository}. All datasets have multivariate quantitative attributes with no missing values. Table \ref{table:Datasets} summarizes these datasets.

\bigskip

Below, we list the machine learning algorithms used as a basis of comparison. Full description of those algorithms is not the scope of the present article and can be found elsewhere in the literature \citep[e.g.,][]{Hastie2009ElemStat}. However, we emphasize how we call these algorithms in each environment (Matlab and OpenCV) to establish common settings for comparison. It is important to note that some options available in Matlab are not available in OpenCV. However, we set the options and call these algorithms in both environment in a way that makes sure they both conform and have common settings; (see Appendix \ref{SecAppendix}.)

\subsection{Classification and Regression Trees (CART)}
CART \citep{Breiman1984ClassificationAnd} is a nonparametric technique that can select the most important variables to predict the output. A tree is constructed from some questions that recursively split the learning sample into parts till it hits a stopping condition.

We use the classification tree by disabling \emph{pruning, surrogate, cross validation} options, so that each leaf node can hold one observation. It is clear that this is an overfitting training; however, we are not interested in building a predictive model here. Rather, we are interested in comparing the execution speed. The impurity measurement used is GINI index. In Matlab, we set the merging leaves option \texttt{mergeleaves} to \texttt{off} and in OpenCV we set the maximum depth per tree as max integer value for signed numbers.

\subsection{Naive Bayes}

It stems from Bayes' theorem, by assuming that features are independent. Naive Bayes is a very simple classifier; however, it outperforms more sophisticated classifiers in some problems. It is more efficient when the dimensionality of the features is high.

There is no additional properties to set for Naive Bayes. We just pass the datasets to both Matlab and OpenCV.

\subsection{Boosting}

Boosting \citep{Freund2004EffBoosting} is one of the most efficient classification algorithms over the past few years. It is an algorithm that combine weak classifiers to produce a strong one. It was originally designed for classification problems but it extends to cover regression problems as well; see \citet{Hastie2009ElemStat,Duda2001PatternClassification}.

We used a CART, with 10 observations per node and  GINI index for impurity measurement, as a weak classifier. The algorithm used for boosting is AdaBoostM1, which is one of the most popular boosting algorithms. Number of ensemble learning cycles is set to 100.

\subsection{Random Forest}

Random Forest \citep{Breiman2001RF} is an ensemble classifier, comprised of many classification trees built on different bootstrap replications from the original dataset \citep{Hastie2009ElemStat}. To classify an object, each tree gives a vote for a class; the object is classified as the majority vote of the trees.

We used 20 trees in the forest, each tree with  minimum of ten observations per node. In both environments, we set \emph{calculating variable importance to on}, and the number of splitting features $m$ to the default value $\sqrt(p)$, where $p$ is the number of dimensions. In Matlab, the trees grow without pruning and we set \texttt{mergeleaves} to \texttt{off}. In OpenCV we set the maximum depth per tree as max integer value for signed numbers.

\subsection{$K$-Nearest Neighbor (KNN)}

KNN is one of the simplest, yet most efficient, classifiers. It classifies the object based on the majority vote of the closest observations (neighbors).

The number of neighbors $K$ is set to one. We used \emph{KD search tree} as our searching algorithm. The distance measure is Euclidean Distance.

\section{Results and Conclusion}\label{SecResults}

\begin{figure*}\centering
    \includegraphics[angle=90,width=\textwidth]{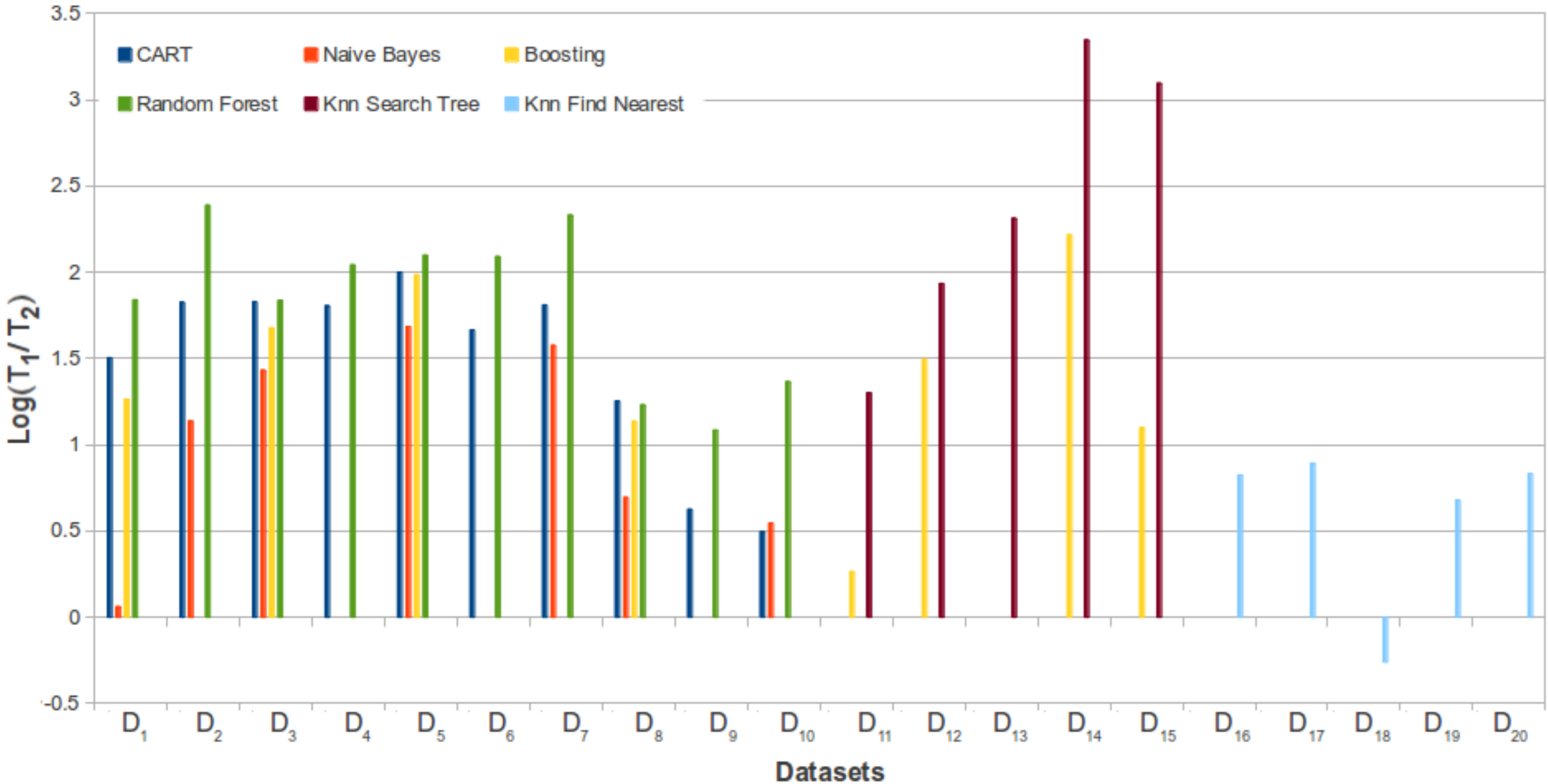}
    \caption{The Log of the execution time ratio ($\log(T_1/T_2)$) between Matlab and OpenCV for
      different Machine Learning algorithms. The $X$-axis is real dataset label as illustrated in
      Table \ref{table:Datasets}.}\label{fig:Results}
\end{figure*}

To compare the speed of Matlab and OpenCV for a particular machine learning algorithm, we run that
algorithm on one of the datasets in Figure \ref{table:Datasets}, using the settings and calling
parameters of Appendix \ref{SecAppendix}. For each dataset, we run the algorithm 1000 times and take
the average of the execution times. Averaging over 1000 experiments is more than necessary since
convergence is reached after few hundreds. Figure \ref{fig:Results} is a plot for the log of the
execution times ratio between Matlab and OpenCV. Each bar represents an algorithm tested on a
dataset. Due to limitation of memory size and CPU performance, not all algorithms are tested on all
datasets.

It is obvious that OpenCV outperforms Matlab across all experiments (with a speed factor up to 80 in some cases), except for the KNN algorithm only on \emph{Pen Based recognition} dataset. It seems that the reason is a combination of number of dimensionality, sample size, and the use of training set. For instance, KNN on datasets $D_{16}$ and $D_{17}$ produced a log time ratio of 0.8 and 0.9 respectively (see Figure). However, when we used only 16 dimensions out of the 10,000 (or 500) dimensions of $D_{16}$ (or $D_{17}$, the ratio was reduced to 0.5 (or 0.06), which means that Matlab became closer to OpenCV. However, two other factors affect these results: the training set and the number of observations. Full understanding for the relative behavior of Matlab and OpenCV in KNN requires more future investigation.

\bigskip

Matlab is a tremendously successful scientific computing environment that helps in developing code in an easy and lucid way. However, when the execution time is an important factor one may need to sacrifice development convenience and write algorithms in a more native way, e.g., by using C, C++, OpenCV, or any similar programming languages.

\bigskip

It is always helpful to distinguish between two purposes of scientific computing: research and product development. For research, one may rely heavily on Matlab, or any similar library-based environment, to experiment new approaches or algorithms. Its rich library and toolboxes, along with data analysis and visualization capabilities, help in rapid research and experimentation and saves the time of implementation from scratch. On the other hand, one of the major important features of a final product is its execution speed that can be achieved by implementation in a native programming language. A prudent trade off between both approaches is achieved by experimenting  using a rich library-based environment; then after settling on a final algorithm or method one can hard-code and deploy it using a more native language.

\section{Acknowledgement}
The first author is grateful to Dr. Ayman Attia, of Helwan University, for fruitful discussions and Omar Elshenawy for technical assistance and early manuscript review.


\section{Appendix: Parameters and Functions Call}\label{SecAppendix}

\subsection{Classification trees}
\subsubsection*{Matlab}
\begin{lstlisting}[style=custommatlab]
classregtree(Data, Target, 'method', 'classification', 'minparent', 1, 'prune', 'off', 'mergeleaves', 'off');
\end{lstlisting}

\subsubsection*{OpenCV}
\begin{lstlisting}[style=customc]
CvDTreeParams Params = CvDTreeParams(Int_Max, 1, 0, false, 100, 0, false, false, NULL);
CvDTree* C_Tree = new CvDTree;
C_Tree->train(Data, CV_ROW_SAMPLE, responses, NULL, NULL, var_type, NULL, Params);
\end{lstlisting}

\subsection{Naive Bayes}

\subsubsection*{Matlab}
\begin{lstlisting}[style=custommatlab]
NaiveBayes.fit(Data, Target);
\end{lstlisting}

\subsubsection*{OpenCV}
\begin{lstlisting}[style=customc]
CvNormalBayesClassifier* Bays= new CvNormalBayesClassifier;
Bays->train(data, responses, 0, 0);
\end{lstlisting}

\subsection{Boosting}

\subsubsection*{Matlab}
\begin{lstlisting}[style=custommatlab]
Tree = ClassificationTree.template('MinParent', 10);
ens = fitensemble(Data, Target, 'AdaBoostM1', 100, Tree, 'type', 'classification');
\end{lstlisting}

\subsubsection*{OpenCV}
\begin{lstlisting}[style=customc]
CvBoost* Boost;
CvBoostParams Params =CvBoostParams(Boost->REAL, 100, 0, Int_Max, false, NULL);
Params.max_categories=100;
Params.min_sample_count=10;
Params.regression_accuracy=0;
Params.split_criteria=Boost->GINI;
Params.truncate_pruned_tree=false;
Params.use_1se_rule=false;
Boost->train(data, CV_ROW_SAMPLE, responses, 0, 0, var_type, 0, Params, false);
\end{lstlisting}

\subsection{Random Forest}

\subsubsection*{Matlab}
\begin{lstlisting}[style=custommatlab]
TreeBagger(20, Data, Target, 'OOBVarImp', 'on', 'Method', 'classification', 'mergeleaves', 'off', 'prune', 'off', 'NVarToSample', 'All');
\end{lstlisting}

\subsubsection*{OpenCV}
\begin{lstlisting}[style=customc]
CvRTrees* RTrees;
CvRTParams Params;
Params = CvRTParams(int_Max, 10, 0, false, 100, 0, true, 0, 20, 0, CV_TERMCRIT_ITER);
RTrees->train(Data, CV_ROW_SAMPLE, Target, 0, 0, var_type , 0, Params);
\end{lstlisting}

\subsection{KNN}

\subsubsection*{Matlab}
\begin{lstlisting}[style=custommatlab]
Tree = createns(Data, 'NSMethod', 'kdtree');
IDX  = knnsearch(Tree, TestData, 'K', 1);
\end{lstlisting}

\subsubsection*{OpenCV}
\begin{lstlisting}[style=customc]
CvKNearest* Knn = new CvKNearest;
Knn->train(data, responses, 0, false, 1);
CvMat* Sample = cvCreateMat(1, Data->cols, CV_32F);
for(j=0;j<Data->rows;j++){
  for(k=0;k<Data->cols;k++)
    cvmSet(Sample, 0, k, cvmGet(Data, j, k));
  Knn->find_nearest(TempData, 1, Results, 0, 0, 0);
}
\end{lstlisting}

\bibliographystyle{elsarticle-harv}
\bibliography{aNEW,publications,booksIhave}

\begin{thebibliography}{27}
\expandafter\ifx\csname natexlab\endcsname\relax\def\natexlab#1{#1}\fi
\providecommand{\url}[1]{\texttt{#1}}
\providecommand{\href}[2]{#2}
\providecommand{\path}[1]{#1}
\providecommand{\DOIprefix}{doi:}
\providecommand{\ArXivprefix}{arXiv:}
\providecommand{\URLprefix}{URL: }
\providecommand{\Pubmedprefix}{pmid:}
\providecommand{\doi}[1]{\href{http://dx.doi.org/#1}{\path{#1}}}
\providecommand{\Pubmed}[1]{\href{pmid:#1}{\path{#1}}}
\providecommand{\bibinfo}[2]{#2}
\ifx\xfnm\relax \def\xfnm[#1]{\unskip,\space#1}\fi
\bibitem[{{Abdel Razek} et~al.(2012){Abdel Razek}, Yousef and
  Mustafa}]{AbdelRazek2012MicroclacificationLIBCAD}
\bibinfo{author}{{Abdel Razek}, N.M.}, \bibinfo{author}{Yousef, W.A.},
  \bibinfo{author}{Mustafa, W.A.}, \bibinfo{year}{2012}.
\newblock \bibinfo{title}{Microcalcification detection with and without
  prototype cad system (libcad): a comparative study}, in:
  \bibinfo{booktitle}{European Society of Radiology}, \bibinfo{publisher}{ECR
  2012 / C-1063}. pp. \bibinfo{pages}{1--15}.
\newblock \URLprefix \url{https://doi.org/10.1594/ecr2012/C-1063},
  \DOIprefix\doi{10.1594/ecr2012/C-1063}.
\bibitem[{{Abdel Razek} et~al.(2013){Abdel Razek}, Yousef and
  Mustafa}]{AbdelRazek2013MicroclacificationLIBCAD}
\bibinfo{author}{{Abdel Razek}, N.M.}, \bibinfo{author}{Yousef, W.A.},
  \bibinfo{author}{Mustafa, W.A.}, \bibinfo{year}{2013}.
\newblock \bibinfo{title}{{Microcalcification Detection With and Without Cad
  System (LIBCAD): a Comparative study}}.
\newblock \bibinfo{journal}{The Egyptian Journal of Radiology and Nuclear
  Medicine} \bibinfo{volume}{44}, \bibinfo{pages}{397--404}.
\bibitem[{Bradski(2000)}]{Bradski2000OpenCV}
\bibinfo{author}{Bradski, G.}, \bibinfo{year}{2000}.
\newblock \bibinfo{title}{{The OpenCV Library}}.
\newblock \bibinfo{journal}{Dr. Dobb's Journal of Software Tools} .
\bibitem[{Breiman(2001)}]{Breiman2001RF}
\bibinfo{author}{Breiman, L.}, \bibinfo{year}{2001}.
\newblock \bibinfo{title}{{Random forests}}.
\newblock \bibinfo{journal}{Machine Learning} \bibinfo{volume}{45},
  \bibinfo{pages}{5--32}.
\newblock \URLprefix \url{https://doi.org/10.1023/a:1010933404324},
  \DOIprefix\doi{10.1023/a:1010933404324}.
\bibitem[{Breiman et~al.(1984)Breiman, Friedman, Olshen and
  Stone}]{Breiman1984ClassificationAnd}
\bibinfo{author}{Breiman, L.}, \bibinfo{author}{Friedman, J.},
  \bibinfo{author}{Olshen, R.}, \bibinfo{author}{Stone, C.},
  \bibinfo{year}{1984}.
\newblock \bibinfo{title}{Classification and regression trees}.
\newblock The Wadsworth statistics/probability series,
  \bibinfo{publisher}{Wadsworth International Group},
  \bibinfo{address}{Belmont, Calif.}
\bibitem[{Chen et~al.(2012a)Chen, Gallas and Yousef}]{Chen2012ClassVar}
\bibinfo{author}{Chen, W.}, \bibinfo{author}{Gallas, B.D.},
  \bibinfo{author}{Yousef, W.A.}, \bibinfo{year}{2012}a.
\newblock \bibinfo{title}{{Classifier Variability: Accounting for Training and
  testing}}.
\newblock \bibinfo{journal}{Pattern Recognition} \bibinfo{volume}{45},
  \bibinfo{pages}{2661--2671}.
\newblock \URLprefix \url{https://doi.org/10.1016/j.patcog.2011.12.024},
  \DOIprefix\doi{10.1016/j.patcog.2011.12.024}.
\bibitem[{Chen et~al.(2012b)Chen, Yousef, Gallas, Hsu, Lababidi, Tang,
  Pennello, Symmans and Pusztai}]{Chen2012UncertEst}
\bibinfo{author}{Chen, W.}, \bibinfo{author}{Yousef, W.A.},
  \bibinfo{author}{Gallas, B.D.}, \bibinfo{author}{Hsu, E.R.},
  \bibinfo{author}{Lababidi, S.}, \bibinfo{author}{Tang, R.},
  \bibinfo{author}{Pennello, G.A.}, \bibinfo{author}{Symmans, W.F.},
  \bibinfo{author}{Pusztai, L.}, \bibinfo{year}{2012}b.
\newblock \bibinfo{title}{{Uncertainty Estimation With a Finite Dataset in the
  Assessment of Classification models}}.
\newblock \bibinfo{journal}{Computational Statistics {\&} Data Analysis}
  \bibinfo{volume}{56}, \bibinfo{pages}{1016--1027}.
\newblock \DOIprefix\doi{10.1016/j.csda.2011.05.024}.
\bibitem[{Duda et~al.(2001)Duda, Hart and
  Stork}]{Duda2001PatternClassification}
\bibinfo{author}{Duda, R.O.}, \bibinfo{author}{Hart, P.E.},
  \bibinfo{author}{Stork, D.G.}, \bibinfo{year}{2001}.
\newblock \bibinfo{title}{{Pattern classification}}.
\newblock \bibinfo{edition}{2nd} ed., \bibinfo{publisher}{Wiley},
  \bibinfo{address}{New York}.
\bibitem[{Freund et~al.(2004)Freund, Iyer, Schapire and
  Singer}]{Freund2004EffBoosting}
\bibinfo{author}{Freund, Y.}, \bibinfo{author}{Iyer, R.},
  \bibinfo{author}{Schapire, R.E.}, \bibinfo{author}{Singer, Y.},
  \bibinfo{year}{2004}.
\newblock \bibinfo{title}{{An Efficient Boosting Algorithm for Combining
  preferences}}.
\newblock \bibinfo{journal}{Journal of Machine Learning Research}
  \bibinfo{volume}{4}, \bibinfo{pages}{933--969}.
\bibitem[{Hastie et~al.(2009)Hastie, Tibshirani and
  Friedman}]{Hastie2009ElemStat}
\bibinfo{author}{Hastie, T.}, \bibinfo{author}{Tibshirani, R.},
  \bibinfo{author}{Friedman, J.H.}, \bibinfo{year}{2009}.
\newblock \bibinfo{title}{{The elements of statistical learning: data mining,
  inference, and prediction}}.
\newblock \bibinfo{edition}{2nd} ed., \bibinfo{publisher}{Springer},
  \bibinfo{address}{New York}.
\bibitem[{Intel(2011)}]{Intel2011TBB}
\bibinfo{author}{Intel}, \bibinfo{year}{2011}.
\newblock \bibinfo{title}{Intel® threading building blocks}.
\newblock \URLprefix \url{http://threadingbuildingblocks.org/}.
\bibitem[{MATLAB(2011)}]{MathWorks2011MATLAB}
\bibinfo{author}{MATLAB}, \bibinfo{year}{2011}.
\newblock \bibinfo{title}{version 7.12.0.635 (R2011a)}.
\newblock \bibinfo{publisher}{The MathWorks Inc.}, \bibinfo{address}{Natick,
  Massachusetts}.
\bibitem[{Newman and Asuncion(2007)}]{Blake1998UCIRepository}
\bibinfo{author}{Newman, D.J.}, \bibinfo{author}{Asuncion, A.},
  \bibinfo{year}{2007}.
\newblock \bibinfo{title}{{{UCI} Machine Learning Repository}}.
\newblock \bibinfo{journal}{University of California, Irvine, Dept. of
  Information and Computer Sciences} .
\bibitem[{Shi et~al.(2010)Shi, Campbell, Jones, Campagne, Wen, Walker, Su, Chu,
  Goodsaid, Pusztai, {Shaughnessy Jr.}, Oberthuer, Thomas, Paules, Fielden,
  Barlogie, Chen, Du, Fischer, Furlanello, Gallas, Ge, Megherbi, Symmans, Wang,
  Zhang, Bitter, Brors, Bushel, Bylesjo, Chen, Cheng, Chou, Davison, Delorenzi,
  Deng, Devanarayan, Dix, Dopazo, Dorff, Elloumi, Fan, Fan, Fan, Fang,
  Gonzaludo, Hess, Hong, Huan, Irizarry, Judson, Juraeva, Lababidi, Lambert,
  Li, Li, Li, Lin, Liu, Lobenhofer, Luo, Luo, McCall, Nikolsky, Pennello,
  Perkins, Philip, Popovici, Price, Qian, Scherer, Shi, Shi, Sung,
  Thierry-Mieg, Thierry-Mieg, Thodima, Trygg, Vishnuvajjala, Wang, Wu, Wu, Xie,
  Yousef, Zhang, Zhang, Zhong, Zhou, Zhu, Arasappan, Bao, Lucas, Berthold,
  Brennan, Buness, Catalano, Chang, Chen, Cheng, Cui, Czika, Demichelis, Deng,
  Dosymbekov, Eils, Feng, Fostel, Fulmer-Smentek, Fuscoe, Gatto, Ge, Goldstein,
  Guo, Halbert, Han, Harris, Hatzis, Herman, Huang, Jensen, Jiang, Johnson,
  Jurman, Kahlert, Khuder, Kohl, Li, Li, Li, Li, Liu, Liu, Liu, Meng, Madera,
  Martinez-Murillo, Medina, Meehan, Miclaus, Moffitt, Montaner, Mukherjee,
  Mulligan, Neville, Nikolskaya, Ning, Page, Parker, Parry, Peng, Peterson,
  Phan, Quanz, Ren, Riccadonna, Roter, Samuelson, Schumacher, Shambaugh, Shi,
  Shippy, Si, Smalter, Sotiriou, Soukup, Staedtler, Steiner, Stokes, Sun, Tan,
  Tang, Tezak, Thorn, Tsyganova, Turpaz, Vega, Visintainer, von Frese, Wang,
  Wang, Wang, Wang, Westermann, Willey, Woods, Wu, Xiao, Xu, Xu, Yang, Zeng,
  Zhang, Zhao, Puri, Scherf, Tong and Wolfinger}]{Shi2010MAQCII}
\bibinfo{author}{Shi, L.}, \bibinfo{author}{Campbell, G.},
  \bibinfo{author}{Jones, W.D.}, \bibinfo{author}{Campagne, F.},
  \bibinfo{author}{Wen, Z.}, \bibinfo{author}{Walker, S.J.},
  \bibinfo{author}{Su, Z.}, \bibinfo{author}{Chu, T.M.},
  \bibinfo{author}{Goodsaid, F.M.}, \bibinfo{author}{Pusztai, L.},
  \bibinfo{author}{{Shaughnessy Jr.}, J.D.}, \bibinfo{author}{Oberthuer, A.},
  \bibinfo{author}{Thomas, R.S.}, \bibinfo{author}{Paules, R.S.},
  \bibinfo{author}{Fielden, M.}, \bibinfo{author}{Barlogie, B.},
  \bibinfo{author}{Chen, W.}, \bibinfo{author}{Du, P.},
  \bibinfo{author}{Fischer, M.}, \bibinfo{author}{Furlanello, C.},
  \bibinfo{author}{Gallas, B.D.}, \bibinfo{author}{Ge, X.},
  \bibinfo{author}{Megherbi, D.B.}, \bibinfo{author}{Symmans, W.F.},
  \bibinfo{author}{Wang, M.D.}, \bibinfo{author}{Zhang, J.},
  \bibinfo{author}{Bitter, H.}, \bibinfo{author}{Brors, B.},
  \bibinfo{author}{Bushel, P.R.}, \bibinfo{author}{Bylesjo, M.},
  \bibinfo{author}{Chen, M.}, \bibinfo{author}{Cheng, J.},
  \bibinfo{author}{Chou, J.}, \bibinfo{author}{Davison, T.S.},
  \bibinfo{author}{Delorenzi, M.}, \bibinfo{author}{Deng, Y.},
  \bibinfo{author}{Devanarayan, V.}, \bibinfo{author}{Dix, D.J.},
  \bibinfo{author}{Dopazo, J.}, \bibinfo{author}{Dorff, K.C.},
  \bibinfo{author}{Elloumi, F.}, \bibinfo{author}{Fan, J.},
  \bibinfo{author}{Fan, S.}, \bibinfo{author}{Fan, X.}, \bibinfo{author}{Fang,
  H.}, \bibinfo{author}{Gonzaludo, N.}, \bibinfo{author}{Hess, K.R.},
  \bibinfo{author}{Hong, H.}, \bibinfo{author}{Huan, J.},
  \bibinfo{author}{Irizarry, R.A.}, \bibinfo{author}{Judson, R.},
  \bibinfo{author}{Juraeva, D.}, \bibinfo{author}{Lababidi, S.},
  \bibinfo{author}{Lambert, C.G.}, \bibinfo{author}{Li, L.},
  \bibinfo{author}{Li, Y.}, \bibinfo{author}{Li, Z.}, \bibinfo{author}{Lin,
  S.M.}, \bibinfo{author}{Liu, G.}, \bibinfo{author}{Lobenhofer, E.K.},
  \bibinfo{author}{Luo, J.}, \bibinfo{author}{Luo, W.},
  \bibinfo{author}{McCall, M.N.}, \bibinfo{author}{Nikolsky, Y.},
  \bibinfo{author}{Pennello, G.A.}, \bibinfo{author}{Perkins, R.G.},
  \bibinfo{author}{Philip, R.}, \bibinfo{author}{Popovici, V.},
  \bibinfo{author}{Price, N.D.}, \bibinfo{author}{Qian, F.},
  \bibinfo{author}{Scherer, A.}, \bibinfo{author}{Shi, T.},
  \bibinfo{author}{Shi, W.}, \bibinfo{author}{Sung, J.},
  \bibinfo{author}{Thierry-Mieg, D.}, \bibinfo{author}{Thierry-Mieg, J.},
  \bibinfo{author}{Thodima, V.}, \bibinfo{author}{Trygg, J.},
  \bibinfo{author}{Vishnuvajjala, L.}, \bibinfo{author}{Wang, S.J.},
  \bibinfo{author}{Wu, J.}, \bibinfo{author}{Wu, Y.}, \bibinfo{author}{Xie,
  Q.}, \bibinfo{author}{Yousef, W.A.}, \bibinfo{author}{Zhang, L.},
  \bibinfo{author}{Zhang, X.}, \bibinfo{author}{Zhong, S.},
  \bibinfo{author}{Zhou, Y.}, \bibinfo{author}{Zhu, S.},
  \bibinfo{author}{Arasappan, D.}, \bibinfo{author}{Bao, W.},
  \bibinfo{author}{Lucas, A.B.}, \bibinfo{author}{Berthold, F.},
  \bibinfo{author}{Brennan, R.J.}, \bibinfo{author}{Buness, A.},
  \bibinfo{author}{Catalano, J.G.}, \bibinfo{author}{Chang, C.},
  \bibinfo{author}{Chen, R.}, \bibinfo{author}{Cheng, Y.},
  \bibinfo{author}{Cui, J.}, \bibinfo{author}{Czika, W.},
  \bibinfo{author}{Demichelis, F.}, \bibinfo{author}{Deng, X.},
  \bibinfo{author}{Dosymbekov, D.}, \bibinfo{author}{Eils, R.},
  \bibinfo{author}{Feng, Y.}, \bibinfo{author}{Fostel, J.},
  \bibinfo{author}{Fulmer-Smentek, S.}, \bibinfo{author}{Fuscoe, J.C.},
  \bibinfo{author}{Gatto, L.}, \bibinfo{author}{Ge, W.},
  \bibinfo{author}{Goldstein, D.R.}, \bibinfo{author}{Guo, L.},
  \bibinfo{author}{Halbert, D.N.}, \bibinfo{author}{Han, J.},
  \bibinfo{author}{Harris, S.C.}, \bibinfo{author}{Hatzis, C.},
  \bibinfo{author}{Herman, D.}, \bibinfo{author}{Huang, J.},
  \bibinfo{author}{Jensen, R.V.}, \bibinfo{author}{Jiang, R.},
  \bibinfo{author}{Johnson, C.D.}, \bibinfo{author}{Jurman, G.},
  \bibinfo{author}{Kahlert, Y.}, \bibinfo{author}{Khuder, S.A.},
  \bibinfo{author}{Kohl, M.}, \bibinfo{author}{Li, J.}, \bibinfo{author}{Li,
  M.}, \bibinfo{author}{Li, Q.Z.}, \bibinfo{author}{Li, S.},
  \bibinfo{author}{Liu, J.}, \bibinfo{author}{Liu, Y.}, \bibinfo{author}{Liu,
  Z.}, \bibinfo{author}{Meng, L.}, \bibinfo{author}{Madera, M.},
  \bibinfo{author}{Martinez-Murillo, F.}, \bibinfo{author}{Medina, I.},
  \bibinfo{author}{Meehan, J.}, \bibinfo{author}{Miclaus, K.},
  \bibinfo{author}{Moffitt, R.A.}, \bibinfo{author}{Montaner, D.},
  \bibinfo{author}{Mukherjee, P.}, \bibinfo{author}{Mulligan, G.J.},
  \bibinfo{author}{Neville, P.}, \bibinfo{author}{Nikolskaya, T.},
  \bibinfo{author}{Ning, B.}, \bibinfo{author}{Page, G.P.},
  \bibinfo{author}{Parker, J.}, \bibinfo{author}{Parry, R.M.},
  \bibinfo{author}{Peng, X.}, \bibinfo{author}{Peterson, R.L.},
  \bibinfo{author}{Phan, J.H.}, \bibinfo{author}{Quanz, B.},
  \bibinfo{author}{Ren, Y.}, \bibinfo{author}{Riccadonna, S.},
  \bibinfo{author}{Roter, A.H.}, \bibinfo{author}{Samuelson, F.W.},
  \bibinfo{author}{Schumacher, M.M.}, \bibinfo{author}{Shambaugh, J.D.},
  \bibinfo{author}{Shi, Q.}, \bibinfo{author}{Shippy, R.}, \bibinfo{author}{Si,
  S.}, \bibinfo{author}{Smalter, A.}, \bibinfo{author}{Sotiriou, C.},
  \bibinfo{author}{Soukup, M.}, \bibinfo{author}{Staedtler, F.},
  \bibinfo{author}{Steiner, G.}, \bibinfo{author}{Stokes, T.H.},
  \bibinfo{author}{Sun, Q.}, \bibinfo{author}{Tan, P.Y.},
  \bibinfo{author}{Tang, R.}, \bibinfo{author}{Tezak, Z.},
  \bibinfo{author}{Thorn, B.}, \bibinfo{author}{Tsyganova, M.},
  \bibinfo{author}{Turpaz, Y.}, \bibinfo{author}{Vega, S.C.},
  \bibinfo{author}{Visintainer, R.}, \bibinfo{author}{von Frese, J.},
  \bibinfo{author}{Wang, C.}, \bibinfo{author}{Wang, E.},
  \bibinfo{author}{Wang, J.}, \bibinfo{author}{Wang, W.},
  \bibinfo{author}{Westermann, F.}, \bibinfo{author}{Willey, J.C.},
  \bibinfo{author}{Woods, M.}, \bibinfo{author}{Wu, S.}, \bibinfo{author}{Xiao,
  N.}, \bibinfo{author}{Xu, J.}, \bibinfo{author}{Xu, L.},
  \bibinfo{author}{Yang, L.}, \bibinfo{author}{Zeng, X.},
  \bibinfo{author}{Zhang, M.}, \bibinfo{author}{Zhao, C.},
  \bibinfo{author}{Puri, R.K.}, \bibinfo{author}{Scherf, U.},
  \bibinfo{author}{Tong, W.}, \bibinfo{author}{Wolfinger, R.D.},
  \bibinfo{year}{2010}.
\newblock \bibinfo{title}{The microarray quality control (maqc)-ii study of
  common practices for the development and validation of microarray-based
  predictive models}.
\newblock \bibinfo{journal}{Nat Biotechnol} \bibinfo{volume}{28},
  \bibinfo{pages}{827--838}.
\bibitem[{WillowGarage(2008)}]{WillowGarage}
\bibinfo{author}{WillowGarage}, \bibinfo{year}{2008} \URLprefix
  \url{http://www.willowgarage.com/}.
\bibitem[{Yousef(2013)}]{Yousef2013PAUC}
\bibinfo{author}{Yousef, W.A.}, \bibinfo{year}{2013}.
\newblock \bibinfo{title}{{Assessing Classifiers in Terms of the Partial Area
  Under the Roc curve}}.
\newblock \bibinfo{journal}{Computational Statistics {\&} Data Analysis}
  \bibinfo{volume}{64}, \bibinfo{pages}{51--70}.
\newblock \URLprefix \url{https://doi.org/10.1016/j.csda.2013.02.032}.
\bibitem[{Yousef(2017)}]{Yousef2017MethodSystemForComputer}
\bibinfo{author}{Yousef, W.A.}, \bibinfo{year}{2017}.
\newblock \bibinfo{title}{Method and system for image analysis to detect
  cancer}.
\newblock \URLprefix
  \url{http://appft.uspto.gov/netacgi/nph-Parser?Sect1=PTO1&Sect2=HITOFF&d=PG01&p=1&u=%2Fnetahtml%2FPTO%2Fsrchnum.html&r=1&f=G&l=50&s1=%2220190019291%22.PGNR.&OS=DN/20190019291&RS=DN/20190019291}.
  \bibinfo{note}{patent pending, US 62/531,219}.
\bibitem[{Yousef(2019a)}]{Yousef2019AUCSmoothness-arxiv}
\bibinfo{author}{Yousef, W.A.}, \bibinfo{year}{2019}a.
\newblock \bibinfo{title}{{AUC}: nonparametric estiamtors and their
  smoothness}.
\newblock \bibinfo{journal}{arXiv preprint arXiv:1907.12851} .
\bibitem[{Yousef(2019b)}]{Yousef2019EstimatingStandardErrorCross-arxiv}
\bibinfo{author}{Yousef, W.A.}, \bibinfo{year}{2019}b.
\newblock \bibinfo{title}{Estimating the standard error of
  cross-validation-based estimators of classification rules performance}.
\newblock \bibinfo{journal}{arXiv preprint arXiv:1908.00325} .
\bibitem[{Yousef(2019c)}]{Yousef2019LeisurelyLookVersionsVariants-arxiv}
\bibinfo{author}{Yousef, W.A.}, \bibinfo{year}{2019}c.
\newblock \bibinfo{title}{A leisurely look at versions and variants of the
  cross validation estimator}.
\newblock \bibinfo{journal}{arXiv preprint arXiv:1907.13413} .
\bibitem[{Yousef(2019d)}]{Yousef2019PrudenceWhenAssumingNormality-arxiv}
\bibinfo{author}{Yousef, W.A.}, \bibinfo{year}{2019}d.
\newblock \bibinfo{title}{Prudence when assuming normality: an advice for
  machine learning practitioners}.
\newblock \bibinfo{journal}{arXiv preprint arXiv:1907.12852} .
\bibitem[{Yousef and Chen(2009)}]{Yousef2009EstCVvariability}
\bibinfo{author}{Yousef, W.A.}, \bibinfo{author}{Chen, W.},
  \bibinfo{year}{2009}.
\newblock \bibinfo{title}{{Estimating Cross-Validation Variability}}, in:
  \bibinfo{booktitle}{Proceedings of the 2009 Joint Statistical Meeting,
  Section on Statistics in Epidemiology.}, pp. \bibinfo{pages}{3318--3326}.
\bibitem[{Yousef and Kundu(2014)}]{Yousef2014LearningAlgo}
\bibinfo{author}{Yousef, W.a.}, \bibinfo{author}{Kundu, S.},
  \bibinfo{year}{2014}.
\newblock \bibinfo{title}{{Learning Algorithms May Perform Worse With
  Increasing Training Set Size: Algorithm-Data incompatibility}}.
\newblock \bibinfo{journal}{Computational Statistics {\&} Data Analysis}
  \bibinfo{volume}{74}, \bibinfo{pages}{181--197}.
\newblock \URLprefix \url{https://doi.org/10.1016/j.csda.2013.05.021},
  \DOIprefix\doi{10.1016/j.csda.2013.05.021}.
\bibitem[{Yousef et~al.(2010)Yousef, Mustafa, Ali, Abdelrazek and
  Farrag}]{Yousef2010OnDetecting}
\bibinfo{author}{Yousef, W.a.}, \bibinfo{author}{Mustafa, W.a.},
  \bibinfo{author}{Ali, A.a.}, \bibinfo{author}{Abdelrazek, N.a.},
  \bibinfo{author}{Farrag, A.M.}, \bibinfo{year}{2010}.
\newblock \bibinfo{title}{{On Detecting Abnormalities in Digital mammography}}.
\newblock \bibinfo{journal}{2010 IEEE 39th Applied Imagery Pattern Recognition
  Workshop (AIPR)} , \bibinfo{pages}{1--7}\URLprefix
  \url{https://doi.org/10.1109/AIPR.2010.5759684},
  \DOIprefix\doi{10.1109/AIPR.2010.5759684}.
\bibitem[{Yousef et~al.(2004)Yousef, Wagner and Loew}]{Yousef2004ComparisonOf}
\bibinfo{author}{Yousef, W.A.}, \bibinfo{author}{Wagner, R.F.},
  \bibinfo{author}{Loew, M.H.}, \bibinfo{year}{2004}.
\newblock \bibinfo{title}{{Comparison of Non-Parametric Methods for Assessing
  Classifier Performance in Terms of {\{}ROC{\}} Parameters}}, in:
  \bibinfo{booktitle}{Applied Imagery Pattern Recognition Workshop, 2004.
  Proceedings. 33rd; IEEE Computer Society}, pp. \bibinfo{pages}{190--195}.
\bibitem[{Yousef et~al.(2005)Yousef, Wagner and Loew}]{Yousef2005EstimatingThe}
\bibinfo{author}{Yousef, W.A.}, \bibinfo{author}{Wagner, R.F.},
  \bibinfo{author}{Loew, M.H.}, \bibinfo{year}{2005}.
\newblock \bibinfo{title}{{Estimating the Uncertainty in the Estimated Mean
  Area Under the {\{}ROC{\}} Curve of a Classifier}}.
\newblock \bibinfo{journal}{Pattern Recognition Letters} \bibinfo{volume}{26},
  \bibinfo{pages}{2600--2610}.
\bibitem[{Yousef et~al.(2006)Yousef, Wagner and Loew}]{Yousef2006AssessClass}
\bibinfo{author}{Yousef, W.A.}, \bibinfo{author}{Wagner, R.F.},
  \bibinfo{author}{Loew, M.H.}, \bibinfo{year}{2006}.
\newblock \bibinfo{title}{{Assessing Classifiers From Two Independent Data Sets
  Using {\{}ROC{\}} Analysis: a Nonparametric Approach}}.
\newblock \bibinfo{journal}{Pattern Analysis and Machine Intelligence, IEEE
  Transactions on} \bibinfo{volume}{28}, \bibinfo{pages}{1809--1817}.

\end{thebibliography}

\end {document}